\documentclass{article}

\pdfoutput=1
\PassOptionsToPackage{numbers, compress}{natbib}

    \usepackage[preprint]{neurips_2022}

\usepackage[utf8]{inputenc} 
\usepackage[T1]{fontenc}    
\usepackage{hyperref}       
\usepackage{url}            
\usepackage{booktabs}       
\usepackage{amsfonts}       
\usepackage{nicefrac}       
\usepackage{microtype}      
\usepackage{xcolor}         
\usepackage{graphicx}
\usepackage{xspace}
\usepackage{amsmath}
\usepackage{wrapfig}

\newcommand{\Ours}[0]{\text{BITS}\xspace}

\title{BITS: Bi-level Imitation for Traffic Simulation}

\author{%
  Danfei Xu\thanks{Equal Contribution} \\
  NVIDIA Research \\
  danfeix@nvidia.com \\
  \And
  Yuxiao Chen$^*$ \\
  NVIDIA Research \\
  yuxiaoc@nvidia.com
  \And
  Boris Ivanovic \\
  NVIDIA Research \\
  bivanovic@nvidia.com
  \And
  Marco Pavone \\
  NVIDIA Research \\
  mpavone@nvidia.com
\\
}

\begin{document}

\setlength{\abovedisplayskip}{1pt}
\setlength{\belowdisplayskip}{1pt}

\maketitle

\begin{abstract}
Simulation is the key to scaling up validation and verification for robotic systems such as autonomous vehicles. Despite advances in high-fidelity physics and sensor simulation, a critical gap remains in simulating realistic \emph{behaviors} of road users. This is because, unlike simulating physics and graphics, devising first principle models for human-like behaviors is generally infeasible. In this work, we take a data-driven approach and propose a method that can learn to generate traffic behaviors from real-world driving logs. The method achieves high sample efficiency and behavior diversity by exploiting the bi-level hierarchy of driving behaviors by decoupling the traffic simulation problem into high-level intent inference and low-level driving behavior imitation. The method also incorporates a planning module to obtain stable long-horizon behaviors. We empirically validate our method, named Bi-level Imitation for Traffic Simulation (BITS), with scenarios from two large-scale driving datasets and show that BITS achieves balanced traffic simulation performance in realism, diversity, and long-horizon stability. We also explore ways to evaluate behavior realism and introduce a suite of evaluation metrics for traffic simulation. Finally, as part of our core contributions, we develop and open source a software tool that unifies data formats across different driving datasets and converts scenes from existing datasets into interactive simulation environments. We include additional results at \url{https://sites.google.com/view/bits2022/home}.
\end{abstract}

\section{Introduction}

Simulation is an integral part of developing effective robotic systems. Simulators allow developers to rapidly verify changes and triage erroneous behaviors before deploying to physical systems. Realistic simulators are especially crucial for autonomous vehicles (AVs), because it is costly and potentially dangerous to test new features and changes directly on the road. Yet, despite advances in physics simulation and high-fidelity sensor simulation, AV developers still primarily rely on large-scale, real-world road testing for validation and verification~\cite{WaymoSafety2021,UberATGSafety2020,NVIDIASafety2021,ArgoSafety2021,ZooxSafety2021,MotionalSafety2021,GMSafety2018}. One critical reason why is that existing simulation platforms do not generate realistic \emph{behaviors} for simulated road users, such as cars and pedestrians, which is difficult because, unlike physics and graphics, it is challenging to design models that generate human-like behaviors from first principles. 

Today's mainstream driving simulators synthesize agent behaviors by either replaying recorded driving logs or implementing heuristics-based controllers. While log replay allows for scenario-specific triaging, it is difficult to validate new features as replayed agents do not react to counterfactual ego motions. On the other hand, heuristics-based controllers are often equipped with simple driving logic and can thus respond to new ego behaviors in a closed-loop manner~\cite{SUMO,Aimsun,Vissim}. While these methods can produce plausible traffic flows, synthesizing diverse and complex driving behaviors such as yielding and cutting-in for a large number of real-world scenarios remains challenging. 

On the other hand, learning-based approaches can ground reactive behavior generation in real-world driving logs. For example, recent works show that trajectory forecasting models trained from large-scale driving logs can accurately infer distributions of future agent trajectories in many challenging scenarios~\cite{WaymoLeaderboard,ArgoverseLeaderboard,LyftLeaderboard,NuscenesLeaderboard}. While these methods excel at predicting realistic trajectories, they are brittle under domain shifts such as new scenes with unseen driving behavior, and the multi-agent nature of traffic simulation may cause a combinatorial explosion in the number of agent states. This challenge is exacerbated when applying prediction approaches to closed-loop behavior simulation over long time horizons, as prediction errors at each step compound over time~\cite{ross2010efficient}, leading to divergence and irreversible failures such as collisions and driving off-road.

In this work, we aim to develop a learning-based traffic simulation model that can generate diverse, stable, and realistic traffic behaviors. Our key insight is two-fold. First, while learning stable long-horizon driving behaviors requires large amounts of data, the problem has a natural bi-level hierarchy that can be exploited to improve learning efficiency. Specifically, we decouple the learning problem into high-level intent inference and low-level goal-conditioned control. Our model leverages a 2D birds-eye-view structure of urban driving and learns to generate a spatial distribution of intended goal locations (Fig.~\ref{fig:model}). A low-level controller policy then generates short segments of goal-conditioned behaviors that move agents towards their inferred goals. In addition, the spatial goal distribution aids in disentangling multi-modal behaviors and generates diverse traffic simulations. At the same time, while such a hierarchical policy greatly improves learning efficiency, agents may still encounter unseen situations that the model alone struggles to handle. To stabilize the long-term behaviors of the model, we augment the policy with a prediction-and-planning module. The module samples likely trajectory segments from the hierarchical policy and selects actions with rule-following cost functions as regularization. This way, the overall framework balances between generating human-like behaviors at in-distribution states and preventing divergences at out-of-distribution states.

We name our method \textbf{\Ours} (\textbf{B}i-level \textbf{I}mitation for \textbf{T}raffic \textbf{S}imulation). We evaluate \Ours on two popular driving log datasets, Lyft Level 5~\cite{HoustonZuidhofEtAl2020} and nuScenes~\cite{CaesarBankitiEtAl2019}. The Lyft dataset contains 1000 hours of driving data collected along a 6.8 mile route in Palo Alto. While densely covered, the trajectory annotations are auto-labeled using a perception stack, resulting in abundant labeling errors~\cite{IvanovicLeeEtAl2021}. In contrast, the nuScenes dataset contains 5.5 hours of manually-annotated trajectories spanning two cities (Boston and Singapore) with more diverse scenarios. Through these two datasets, we demonstrate the capability of our method under different types of learning challenges. Beyond generating realistic traffic behaviors, we also explore ways to evaluate the generated behaviors through a suite of analytical and learned metrics, since conventional trajectory generation metrics such as ADE and FDE are ill-fitted for evaluating closed-loop traffic simulation. Finally, as part of our core contributions, we develop and open source a software tool that unifies data formats across different AV datasets and allows users to transform scenarios from existing datasets into interactive simulation environments. We hope that our novel traffic behavior simulation method, along with the evaluation protocol and data interface software, can serve as a foundation for further research on this topic.

\section{Related work}

\textbf{Traffic simulation.}
Approaches for traffic simulation
can broadly be categorized into two groups:
\textit{macroscopic}, studying large scale traffic flows without instantiating individual agent states, and \textit{microscopic}, focusing on individual agents in traffic and simulating their behaviors on the road \cite{traffic_sim_review2009,traffic_sim_review2017}. Since our ultimate goal is to validate autonomous vehicles and their interaction with surrounding agents, we focus our attention on microscopic simulation. Existing systems for microscopic simulation such as SUMO \cite{SUMO}, Aimsun \cite{Aimsun}, and VISSIM \cite{Vissim} use analytical models to control agents in a scene, including cellular automata, the intelligent driver model (IDM), and the optimal velocity model (see \cite{brockfeld2003toward} for a comprehensive review). These analytic models typically have fixed routes for vehicles to follow and separate longitudinal and lateral motions of agents, with some models omitting lateral motion entirely. Accordingly, these analytical microscopic traffic simulation tools lack sufficient complexity and expressiveness for developing or evaluating autonomous driving features. 

\begin{figure}
    \centering
    \includegraphics[width=\linewidth]{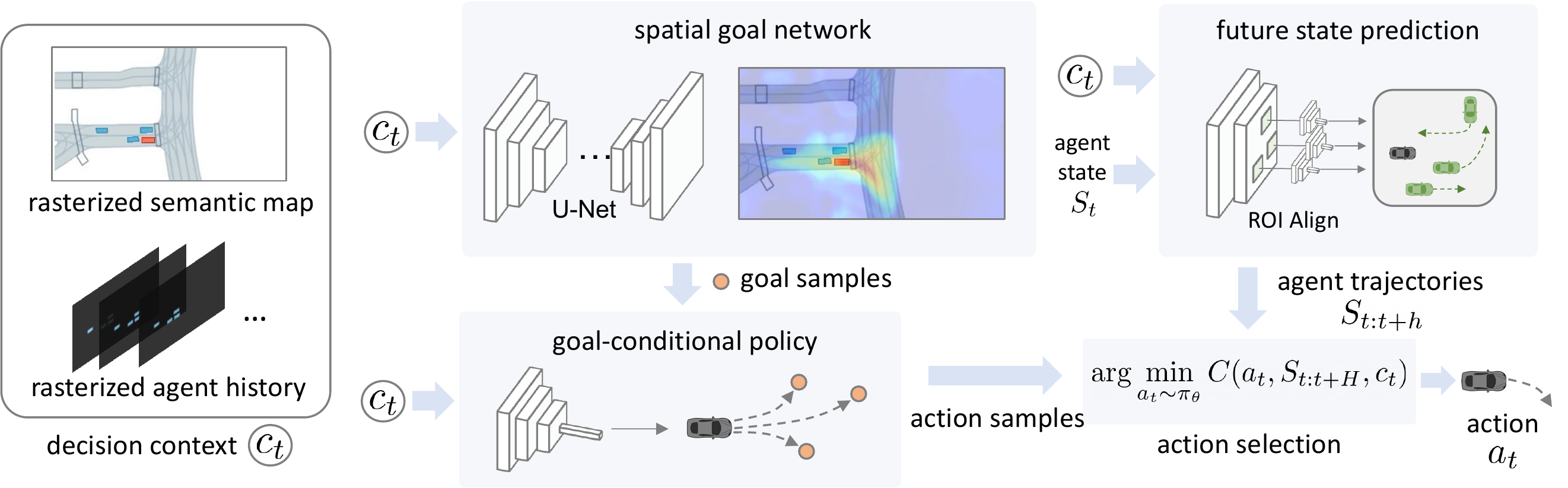}
    \caption{\textbf{\Ours framework overview}: Decision context $c_t$ is a tensor containing the semantic map and rasterized agent history concatenated channel-wise. Given $c_t$ as input, (1) the spatial goal network produces a 2D spatial distribution of short-horizon goals, (2) the goal-conditional policy generates a set of actions for each sampled goal, (3) a trajectory forecasting model predicts the future motion of the neighboring agents, and finally (4), based on the predicted future states, the framework selects the set of actions that minimizes a rule-based cost function.}
    \label{fig:model}
    
    \vspace{-0.5cm}
    
\end{figure}

As a result, recent works have started to develop more expressive models for traffic simulation, generally based on neural networks~\cite{bergamini2021simnet,suo2021trafficsim,rempe2022strive}, that learn from real-world trajectory datasets~\cite{waymo_open_dataset,HoustonZuidhofEtAl2020,CaesarBankitiEtAl2019}. Notably, STRIVE~\cite{rempe2022strive} proposes to generate near-collision scenarios by searching in the latent space of a trajectory prediction model. TrafficSim~\cite{suo2021trafficsim} adapts a graph-based trajectory prediction model to perform scene-level traffic simulation.  Compared to STRIVE, which focuses on generating worst-case scenarios and short open-loop trajectories, we aim to synthesize a broad range of long-horizon closed-loop traffic behaviors. Compared to TrafficSim, which relies on scene-level control to ensure consistent interaction among agents (e.g., coordinated collision avoidance), our method enables each agent to act without coordination with others at the model level. Such an agent-centric setup allows our method to be deployed in more practical simulation usecases, where different types of agents (analytical or learned) are mixed together to interact in a scene. Moreover, we also show that our method outperforms TrafficSim in simulation diversity and stability. 

\textbf{Trajectory prediction.}
A separate set of research aims to predict the trajectories of agents in traffic, with initial works applying analytical models such as Social Forces \cite{HelbingMolnar1995,mehran2009abnormal}, Hidden Markov Models \cite{KitaniZiebartEtAl2012} and IDM \cite{TreiberHenneckeEtAl2000b}. 
However, limited by their expressiveness, they fail to scale to the level of complexity required by autonomous driving. As a result, various data-driven models have been proposed, which directly learn to predict agent behaviors from a wealth of demonstrations. Several early examples include Social GAN \cite{GuptaJohnsonEtAl2018}, GAIL \cite{KueflerMortonEtAl2017}, MFP \cite{TangSalakhutdinov2019}, and DESIRE \cite{LeeChoiEtAl2017}. More recent trajectory prediction models can be further categorized into \textit{agent-centric} models, which generate independent predictions for each agent in a scene \cite{SalzmannIvanovicEtAl2020,LeeChoiEtAl2017,ZhaoXuEtAl2019}, and \textit{scene-centric} models, which generate joint predictions for all (or a subset of) agents in a scene \cite{CasasGulinoEtAl2019,CasasGulinoEtAl2020b}. For a comprehensive review of trajectory prediction methods, see \cite{RudenkoPalmieriEtAl2019}. While these methods perform well over short time-horizons (up to 5s), their performance generally degrades over longer time-horizons. To combat this, many recent state-of-the-art methods adopt a multi-stage approach which first predicts agents' goal locations and subsequently links agents' current positions to their inferred goals \cite{ZhaoWildes2019,MangalamGiraseEtAl2020,RhinehartMcAllisterEtAl2019,ChoiMallaEtAl2020,ZhaoGaoEtAl2020,GuSunEtAl2021,GillesSabatiniEtAl2021,GillesSabatiniEtAl2022a,GillesSabatiniEtAl2022b}.
While these methods solely focus on open-loop prediction, we will show in this work that inferring and conditioning on an agent's goal also improves the stability of long-horizon closed-loop simulation.
 
 \textbf{Imitation learning.} 
Our bi-level imitation learning method is heavily inspired by literature in hierarchical decision making and multimodal imitation learning. A hierarchical policy consists of a high-level planner that sets abstract goals and a low-level policy that learns to achieve the goals~\cite{sutton1999between,dayan1992feudal,bacon2017option,vezhnevets2017feudal,le2018hierarchical,shiarlis2018taco,pertsch2020long}. Methods along this vein have favorable properties such as compositionality~\cite{shiarlis2018taco} and interpretability~\cite{shu2017hierarchical}, and they have achieved superior performance in long-horizon tasks especially~\cite{vezhnevets2017feudal,le2018hierarchical,shiarlis2018taco,pertsch2020long}. At the same time, an equally important desideratum of traffic simulation is behavior diversity, which most prior works neglect. Multimodal imitation learning was recently studied in the manipulation domain~\cite{lynch2020learning,mandlekar2020iris,mandlekar2020learning,florence2022implicit}. Notably, GTI~\cite{mandlekar2020learning} trains a CVAE-based planner to set multi-modal subgoals in observation space for a low-level goal-conditioning controller to achieve. Our method adopts a similar hierarchical structure but instead exploits the domain structure of driving to efficiently represent goal distributions as 2D birds-eye view spatial maps (shown in Fig.~\ref{fig:model}). We empirically show that our method generates realistic, diverse, and stable long-horizon traffic simulations.

\section{Bi-level imitation for traffic simulation}

In this section, we dive into the details of the traffic simulation problem and our primary technical contributions. We propose (1) a hierarchical imitation learning framework that generates diverse and realistic traffic behaviors, (2) a prediction-and-planning module that stabilizes long-horizon simulation, and (3) a suite of analytical and learned evaluation metrics for traffic simulation. Specific implementation details (network architectures, shapes) will be described in Sec.~\ref{sec:eval_setup} and the Appendix.

\subsection{Traffic simulation as imitation learning}
We take an agent-centric approach to traffic simulation, i.e., each agent makes decisions in a decentralized manner without explicit coordination. As mentioned previously, this allows for flexible integration with existing simulation frameworks containing other types of simulated agents and encourages the emergence of new interactive behaviors. We focus on simulating vehicle traffic in this work, but an agent can be any type of road user captured in driving logs (e.g., cyclists, pedestrians). We use $s$ and $c$ to denote the dynamic state and decision-relevant context for an agent, respectively. Specifically, state $s$ includes the position, heading, and velocity of an agent. Context $c=(I, S)$ includes a local semantic map $I$ and the $h$ previous states of an agent and its $N$ neighboring agents $S_{t-h:t}=\{s^{(0)}_{t-h:t}, s^{(1)}_{t-h:t}, ..., s^{(N)}_{t-h:t}\}$. Given the decision context information $c_t$ and the current state $s_t$, the goal of a traffic simulation model $\pi_\theta$ is to generate the next state of the agent $s_{t+1} = \mathcal{T}(\pi_\theta(c_t), s_t)$ subject to a dynamics transition function $\mathcal{T}(\cdot)$. We use a simple unicycle model with dynamics constraints as $\mathcal{T}$ and defer more realistic vehicle dynamics models to future works. 

We leverage driving logs captured in the real world~\cite{HoustonZuidhofEtAl2020,CaesarBankitiEtAl2019} to train our traffic model. Since log data readily includes semantic maps and the trajectories of all observed agents, we can treat driving logs as a set of multi-agent expert demonstration sequences $\tau=\{c^{(i)}_0, s^{(i)}_0, c^{(i)}_1, s^{(i)}_1, ..., c^{(i)}_T, s^{(i)}_T\}_{i=0}^N$ and formulate traffic simulation as a supervised imitation learning problem. However, the nature of urban driving poses significant technical challenges. First, the decision process is partially-observed as the model does not have access to the underlying intent of the demonstrator and other decision-relevant cues such as the turn signals of other vehicles. Accordingly, action supervisions are inherently ambiguous and are usually modeled with probabilistic distributions~\cite{SalzmannIvanovicEtAl2020,LeeChoiEtAl2017,ZhaoXuEtAl2019}. Although this ambiguity complicates training, effectively modeling action distributions also enables generating diverse counterfactual traffic simulations. Second, since each agent acts without explicit coordination, their joint behaviors create a combinatorial space of possible future states. Such uncertainty makes generating stable traffic simulations extremely challenging. Below, we describe how our approach can generate multimodal simulations with a stochastic hierarchical policy and mitigate uncertainty in state evolution with a prediction-and-planning module. 

\subsection{Bi-level imitation learning for multi-modal behavior generation}
The goal of our traffic simulation model is to produce diverse and plausible behaviors by learning from real-world driving logs as demonstrations. Most existing methods in trajectory prediction use deep latent variable models (e.g., VAEs) to capture the behavior distributions. However, findings in the imitation learning literature~\cite{RossGordonEtAl2011} suggest that learning to generate stable long-horizon behaviors requires a large amount of training data. Our method instead decomposes the learning problem into (1) training a high-level goal network that captures the spatial distribution of possible short-term goals, and (2) training a deterministic goal-conditional policy that learns to reach the predicted goal. The spatial goal network exploits the 2D birds-eye-view structure of driving motion and represents the spatial goal distribution efficiently with a 2D grid. This decomposition additionally moves the burden of modeling multi-modal trajectories to the high-level goal predictor, enabling the low-level goal-conditioned policy to reuse goal-reaching skills to improve sample efficiency.

\textbf{Spatial goal network.} The spatial goal network is trained to predict the distribution of the short-term goal pose (2D position and heading) $p(\tilde{s}_{t+H}|c_t)$ of an agent given its decision context $c_t$. Following prior works~\cite{bergamini2021simnet,konev2021motioncnn}, we first encode the decision context into a rasterized semantic map, which includes the semantic map $I$ and past agent trajectories rasterized as 2D bounding boxes in additional channels. The model takes as input the rasterized semantic map and outputs a 2D grid of goal likelihood as well as residual components to refine the predicted goal location. The output takes the shape of a 4-channel tensor with the same spatial size as the input rasterized map. Channel $0$ is the likelihood of the coarse goal location 2D probability map. Each pixel in channel $1$ and $2$ is the $(x, y)$ scalar residual relative to the grid location. Channel $3$ is the heading prediction at each grid location. Once a location is selected based on the probability map in channel $0$, the location is corrected by the residual and transformed into a goal pose $\tilde{s}_{t+H}$ in the agent local coordinate frame. We treat the 2D location map as a joint distribution and train via cross-entropy loss across locations. The other channels are trained with masked regression losses (e.g., squared error). 

\textbf{Goal-conditional policy.} The goal-conditional policy takes the form of a deterministic trajectory generator $s_{t:t+H}=\pi_{\theta}(c_t, \tilde{s}_{t+H})$. Although we may further augment the policy with stochastic components, we empirically found it unnecessary as the short-term goal largely reduces the uncertainty in prediction. Inspired by prior works~\cite{SalzmannIvanovicEtAl2020,suo2021trafficsim}, instead of directly regressing each state in an agent's trajectory
the model predicts controls (velocity, change of heading) at each future time step and forward integrates them through an agent's dynamics model (e.g., extended unicycle dynamics~\cite{LaValle2006BetterUnicycle} for vehicles). The errors between the predicted and reference trajectory are then backpropagated directly through the dynamics model. Overall, this strategy provides a strong dynamically-grounded learning signal which corrects predictions at earlier steps to reduce errors at later steps.

\subsection{Prediction and planning for long-horizon stability}\label{sec:pred_and_plan}
So far, we have described a bi-level imitation learning method that can generate plausible traffic simulations from limited data. The policy can synthesize diverse behaviors by sampling from the multi-modal spatial goal predictor. However, the performance of the policy remains bounded by the size and coverage of training data. Driving logs are biased towards nominal behaviors and contain almost no safety-critical situations such as collisions or driving off-road. The objective of generating diverse behaviors further amplifies this challenge, as agents are encouraged to enter previously-unseen regions of the map and create new interactions. As a result, to achieve stable long-horizon simulations, agents must generate reasonable behaviors even at states where guidance from training data is lacking.

To this end, we propose to augment our policy with a \emph{prediction-and-planning} module to stabilize long-horizon rollouts. The module draws action samples $a_t$ from the stochastic bi-level policy $\pi_\theta$ described above and selects the action that minimizes a rule-based cost function $C$ given the predicted future states of the environment $S_{t:t+H}$, that is,  $\arg\min_{a_t\sim \pi_{\theta}} C(a_t, S_{t:t+H}, c_t)$.
This approach is similar to the motion planning pipeline in a typical modular AV stack, with the important difference that we use our learned policy to generate human-like motion trajectory candidates. The key idea is that the policy $\pi_\theta$ can directly follow the data likelihood at in-distribution states, where most action samples are rule-following, and receive corrective guidance at states where the most likely actions may lead to bad consequences. In addition, the sampling module allows for agile adjustment of the simulator (e.g. level of diversity, emphasis on multiple objectives) without retraining. Below we describe the model for future state prediction and the cost function for action selection.

\textbf{Future state prediction.} Since we assume an analytical vehicle dynamics model and known static map, the main task of the model is to predict the future motion trajectories of nearby agents. We follow a typical trajectory prediction pipeline and featurize each agent by its local and global scene context. Specifically, we use RoIAlign~\cite{he2017mask} to crop the features extracted by the intermediate layer of a deep CNN. The per-agent features are then concatenated with a global scene context feature (extracted by the final layer of the same CNN) to make the final trajectory prediction $s^{(i)}_{t:t+H}$ for each neighboring agent $i$. The model is illustrated in Fig.~\ref{fig:model}. We use deterministic prediction in this work and defer more sophisticated probabilistic prediction and planning to future work.

\textbf{Cost-based trajectory selection.} We consider two rule-based costs: collision and road departure. The collision cost is computed based on distances from the corners of two bounding box rectangles using
distances between the four corners. 
To calculate the road departure cost, we first generate a distance map that records the Manhattan distance to the drivable area in pixel space (obtained efficiently via $D$ convolution steps). The resulting distance map assigns zero to points within the drivable area, with values increasing outside the drivable area until saturating at $D$. We directly use this Manhattan distance value as a penalty. Note that both cost terms are zero for nominal trajectories, i.e., trajectories that do not result in collision and road departure, minimizing their effect on selecting among rule-following action samples.

\subsection{Evaluation metrics for traffic simulation}
\label{ssec:metrics}
Designing metrics for simulation is particularly difficult because of the lack of ground truth. As a result, metrics such as average displacement error (ADE) and final displacement error (FDE), commonly used to evaluate trajectory prediction, do not suit the evaluation of simulation models. To address this evaluation gap, we propose three types of simulation metrics: (i) metrics measuring how much simulated agents violate common traffic rules, such as driving off-road or causing collisions with other agents; (ii) metrics measuring the statistics of simulation rollouts, including resemblance to collected driving logs in terms of driving characteristics such as speed profile, control effort, coverage of the driving area, and behavior diversity between different simulation trials; (iii) data-driven metrics learned from real-world driving logs, such as measuring the likelihood of simulation rollouts under data-driven trajectory forecasting models. Here we describe (ii) and (iii) in details.

\textbf{Coverage and diversity.} To calculate how much of a scene is covered by the agents, we first compute a simulator's trajectory distribution via Density Estimation with a Gaussian kernel over all time steps of the simulator's rollouts, focusing on the 2D spatial distribution of the trajectory. To measure the coverage of the map, we count the number of grid points where the KDE estimate is above a threshold, separating the count between drivable areas and non-drivable areas. To measure the diversity of stochastic policies, we run multiple trials with the same initial condition and collect the density estimates for each trial. Given two different trials, we compute the Wasserstein distance between the two density profiles. In particular, all grid points with non-zero density are flattened and a distance matrix is computed containing the Euclidean distances between every pair of grid points. The density profile is then normalized to sum to 1 and the Wasserstein distance can be computed efficiently as in~\cite{pele2009}. For $n$ trials of the same scene, we calculate the Wasserstein distances between the $n(n-1)/2$ pairs of density profiles and take the mean as a metric for diversity:
\begin{equation*}
    \mathrm{Diversity}= \frac{2}{n(n-1)}\sum_{i=1}^{n-1}\sum_{j=i+1}^{n} \mathrm{Wass}(\rho_i,\rho_j),
\end{equation*}
where $\mathrm{Wass(\cdot,\cdot)}$ is the Wasserstein distance and $\rho_i$ is the density profile for the $i$-th trial.

\textbf{Learned metrics.} There are many potential ways to learn a metric that evaluates whether simulated behavior is human-like. One such way is to evaluate simulated agent trajectory likelihoods using a prediction model trained from real-world driving logs. To evaluate this possibility, we develop an occupancy-based prediction model that predicts where an agent will be in future time steps. The model uses a similar structure to our spatial goal network and discretizes the position space into bins. The model is trained to minimize a cross-entropy loss function and its predictions are then used to compute the trajectory likelihoods of simulated trajectories. We include more details in the Appendix.



\section{Evaluation}
Our experiments seek to validate the primary claims that (1) \Ours can generate plausible behaviors by learning from real-world driving logs, (2) compared to other flat policies, our hierarchical policy learning framework achieves better sample efficiency and behavior diversity, (3) our proposed prediction-and-planning module is effective at providing corrective guidance for out-of-distribution states. We conduct evaluations with two large-scale real-world driving datasets, Lyft Level 5~\cite{HoustonZuidhofEtAl2020} and nuScenes~\cite{CaesarBankitiEtAl2019}. Since learning-based traffic simulation is a new topic and lacks a standardized benchmark, in this work we also develop and open source a software framework that unifies data formats across different AV datasets (starting with the two used in this work) and can transform scenes from datasets to interactive simulation environments. We use this framework to run closed-loop simulations and report performance based on the metrics described in Sec.~\ref{ssec:metrics}. 

\subsection{Evaluation setup}\label{sec:eval_setup}
\textbf{Datasets.} The Lyft dataset~\cite{HoustonZuidhofEtAl2020} contains 1000 hours of driving data collected along a 6.8 mile route in Palo Alto. The dataset contains many repeated trips along each road segment. Since the annotations are auto-generated using a perception stack, there are many labeling errors including inaccurate agent positions, headings, and semantic types~\cite{IvanovicLeeEtAl2021}. In contrast, nuScenes~\cite{CaesarBankitiEtAl2019} contains 5.5 hours of accurate manually-labeled trajectories spanning two cities (Boston and Singapore) with more diverse scenarios and denser traffic. Through these datasets, we compare our method to baselines under different learning challenges such as noisy labels and small training sets. For both datasets, we train all models on trajectories from the train split and conduct evaluation on 100 scenes randomly sampled from the validation split. We consider only vehicle simulation in this paper and defer simulating other types of agents (e.g., pedestrians, cyclists) to future works.

\textbf{Simulation environments.} As stated above, we initialize our simulation environments from real driving data, leading to realistic agent placements and dynamic states. Note that all scenes are drawn from the validation split previously unseen to the trained models. Each agent in the scene is independently controlled by replicas of the same model. The simulation runs at a frequency of 10 Hz and results are reported on 20-seconds simulation episodes.

\begin{table}[h]
\caption{Quantitative results on the Lyft Dataset~\cite{HoustonZuidhofEtAl2020}}
\begin{tabular}{l|ccccc|ccc}
\hline
                                         & FR$\downarrow$ & coll$\downarrow$ & offroad$\downarrow$ & coverage$\uparrow$ & diversity$\uparrow$ & speed & jerk & sADE  \\ \hline
SimNet\cite{bergamini2021simnet}       & 38.35          & 35.57            & 1.38                & 460.44             & 0.00                & 1.65  & 3.85 & 3.19  \\
SocialGAN\cite{GuptaJohnsonEtAl2018}        & 64.96          & 42.86            & 19.45               & 189.98             & 9.02                & 1.48  & 5.17 & 13.22 \\
SocialGAN+p                              & 69.41          & 42.96            & 25.12               & 131.47             & 7.64                & 1.51  & 4.78 & 13.90 \\
TPP\cite{SalzmannIvanovicEtAl2020}     & 15.62          & 14.65            & 0.59                & 495.69             & 3.23                & 1.14  & 2.60 & 5.50  \\
TPP+p                                    & 16.03          & 15.12            & 0.65                & 508.16             & 2.75                & 1.23  & 2.32 & 5.44  \\
TrafficSim\cite{suo2021trafficsim}     & 26.98          & 15.98            & 5.76                & 566.35             & 7.68                & 1.50  & 3.82 & 5.89  \\
TrafficSim+p                             & 22.97          & 13.58            & 4.70                & 617.50             & 7.96                & 1.73  & 3.09 & 6.04  \\
\Ours (max)                            & 20.71          & 18.75            & 1.18                & 443.93             & 0.00                & 0.76  & 4.44 & 7.77  \\
\Ours (sample) & 25.37          & 22.37            & 1.27                & 780.52             & 16.84               & 1.86  & 4.29 & 7.78  \\
\Ours          & \textbf{9.97}  & \textbf{8.66}    & \textbf{0.46}       & \textbf{1014.43}   & \textbf{22.94}      & 1.96  & 3.75 & 11.21 \\ \hline
Dataset                                  & 18.36          & 17.49            & 0.84                & 327.25             & 0.00                & 0.00  & 0.00 & 0.00  \\ \hline
\end{tabular}
\label{tab:lyft}
\vspace{-0.2cm}
\end{table}

\begin{table}[h]
\caption{Quantitative results on the nuScenes Dataset~\cite{CaesarBankitiEtAl2019}}
\begin{tabular}{l|ccccc|ccc}
\hline
                                       & FR$\downarrow$ & coll$\downarrow$ & offroad$\downarrow$ & coverage$\uparrow$ & diversity$\uparrow$ & speed & jerk  & sADE  \\ \hline
SimNet\cite{bergamini2021simnet}     & 24.58          & 15.80            & 3.05                & 395.21             & 0.00                & 6.81  & 18.61 & 7.01  \\
SocialGAN\cite{GuptaJohnsonEtAl2018} & 71.33          & 33.81            & 19.71               & 154.21             & 2.29                & 38.44 & 42.51 & 19.89 \\
SocialGAN+p                            & 71.86          & 35.84            & 18.26               & 151.61             & 2.32                & 37.07 & 42.48 & 19.76 \\
TPP\cite{SalzmannIvanovicEtAl2020}   & 49.76          & 33.23            & 8.27                & 489.36             & 7.06                & 8.37  & 4.61  & 12.13 \\
TPP+p                                  & 9.84           & 9.52             & \textbf{0.04}       & 661.07             & 7.73                & 8.17  & 5.96  & 13.09 \\
TrafficSim\cite{suo2021trafficsim}   & 27.08          & 20.39            & 3.42                & 861.00             & 4.28                & 10.03 & 10.01 & 10.51 \\
TrafficSim+p                           & 25.97          & 18.45            & 3.61                & 933.53             & 4.43                & 10.44 & 11.93 & 11.06 \\
\Ours (max)                          & 13.24          & 11.51            & 0.30                & 559.58             & 0.00                & 6.35  & 13.87 & 6.41  \\
\Ours (sample)                       & 14.72          & 13.79            & 0.17                & 888.12             & 6.36                & 6.54  & 13.44 & 6.70  \\
\Ours                                & \textbf{6.63}  & \textbf{5.67}    & 0.14                & \textbf{1122.94}   & \textbf{8.40}       & 6.91  & 13.20 & 8.39  \\ \hline
Dataset                                & 11.99          & 11.39            & 0.42                & 397.92             & 0.00                & 0.00  & 0.00  & 0.00  \\ \hline
\end{tabular}
\label{tab:nusc}
\vspace{-0.5cm}
\end{table}

\textbf{Baselines.} We consider methods from both the traffic simulation and trajectory prediction literature. \textbf{SimNet}~\cite{bergamini2021simnet} is a deterministic behavior-cloning model for traffic simulation. 
\textbf{TrafficSim} is an agent-centric adaptation of the original traffic simulation method in~\cite{suo2021trafficsim} that features an isotropic Gaussian CVAE. We remove the scene consistency loss in training since we do not assume control over all agents. \textbf{SocialGAN}~\cite{GuptaJohnsonEtAl2018} learns to generate trajectories through adversarial imitation. \textbf{TPP} is adapted from Trajectron++~\cite{SalzmannIvanovicEtAl2020}, comprised of a discrete CVAE with Gaussian trajectory decoder for each discrete mode. We also consider variants of these methods augmented with our planning-and-control module (marked with ``\textbf{+p}''), i.e., selecting future action samples with a cost function. We also evaluate ablations of our method, \textbf{\Ours (max)} takes the maximum-likelihood action instead of sampling and \textbf{\Ours(sample)} samples actions for rollouts without the prediction-and-planning module described in Sec.~\ref{sec:pred_and_plan}. All methods share the same rasterized input format, ResNet-18 encoder backbone, and MLP-based trajectory decoder networks.

\textbf{Metrics.} As mentioned in Sec.~\ref{ssec:metrics}, designing evaluation metrics for traffic simulation is challenging as there are not single quantities that can summarize the performance of a method, and we cannot easily compare with ground truth dataset trajectories as our goal is to generate new and diverse simulation rollouts. To address this problem, we consider three types of evaluation metrics.

\emph{Environment metrics} measure rule violations, environment coverage, and trajectory diversity. Both \textbf{coverage} and \textbf{diversity} as described in Sec.~\ref{ssec:metrics} are calculated from 5 simulation trials with different seeds per scene. We define a critical failure as an agent colliding with other agents or driving off-road for more than 1s. Failure Rate (\textbf{FR}) is the average fraction of agents experiencing a critical failure in a scene. We also report raw collision rate (\textbf{coll}) and road departure rate (\textbf{offroad}). 

\emph{Dataset metrics} compare simulation and ground truth data statistics. They are computed using a normalized Wasserstein distance between the histograms of the driving profiles of the simulated and recorded trajectories. We focus on \textbf{speed} and \textbf{jerk}, commonly used as driver comfort metrics, in the main paper and include others in the Appendix. We also report scene Average Distance Error (\textbf{sADE}) which measures the average position differences between simulated and recorded trajectories. Note that sADE is not suitable for measuring simulation realism and is included only as a reference since it heavily penalizes alternative simulations (e.g., turning left vs.~going straight).

Finally, \emph{learned metrics} as described in Sec.~\ref{ssec:metrics} measure simulation realism based on a likelihood model trained from real-world driving log.

\vspace{-0.25cm}

\subsection{Main results}
\begin{figure}
    \centering
    \includegraphics[width=1.0\linewidth]{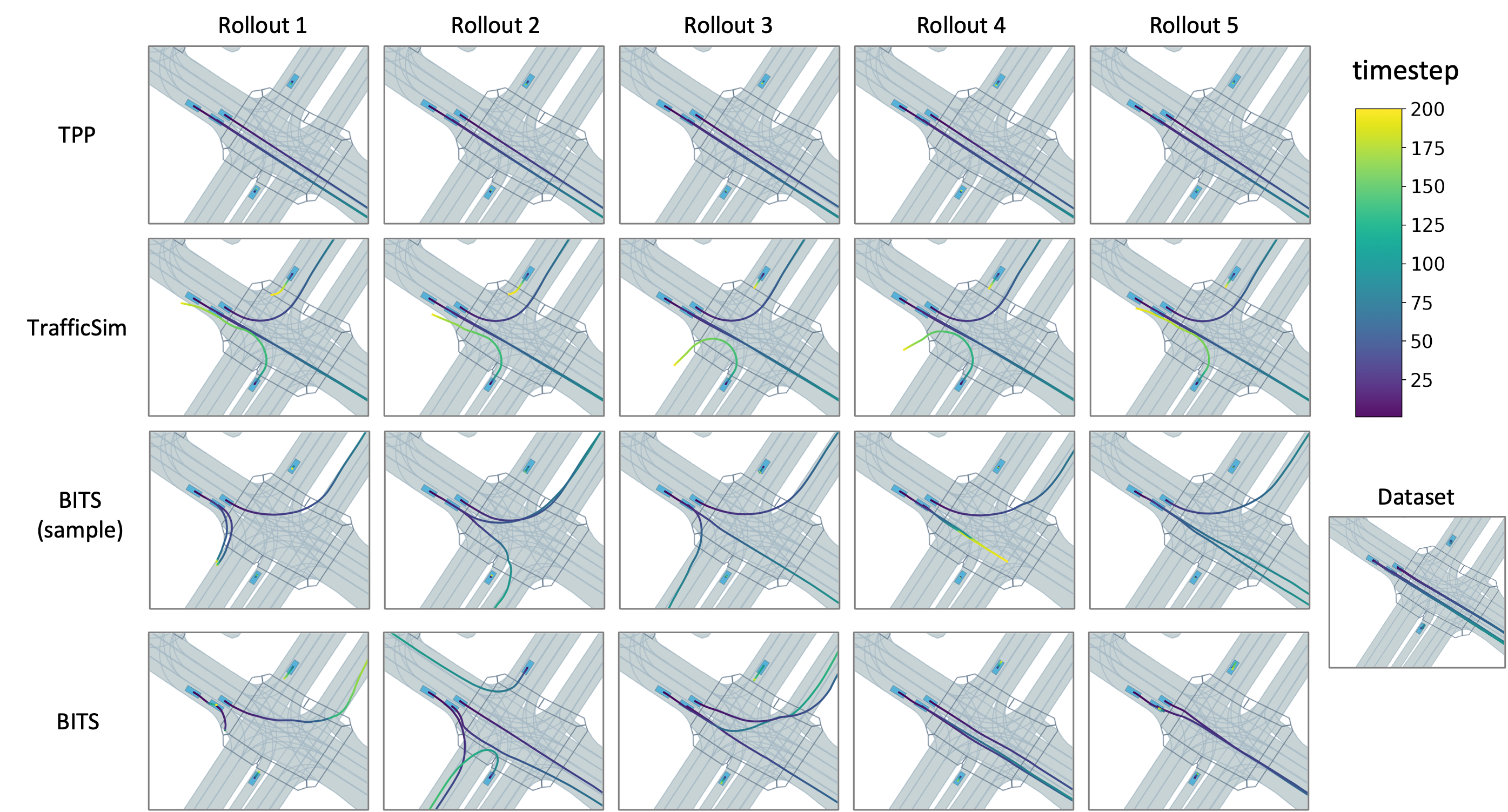}
    \caption{Trajectories generated by each stochastic method over 5 trials in the Lyft dataset~\cite{HoustonZuidhofEtAl2020}. Our method (\Ours, last row) generates diverse and stable long-horizon simulation rollouts (visualized as colored lines emanating from agents). Other methods suffer from a lack of diversity (e.g., TPP~\cite{SalzmannIvanovicEtAl2020} on top) or high collision and off-road rates (e.g., TrafficSim~\cite{suo2021trafficsim} in the second row). Agents are represented with blue bounding boxes and trajectory line color denotes simulation timestep.}
    \label{fig:diversity}
    \vspace{-0.5cm}
\end{figure}

\begin{figure*}
    \centering
    \includegraphics[width=0.6\linewidth]{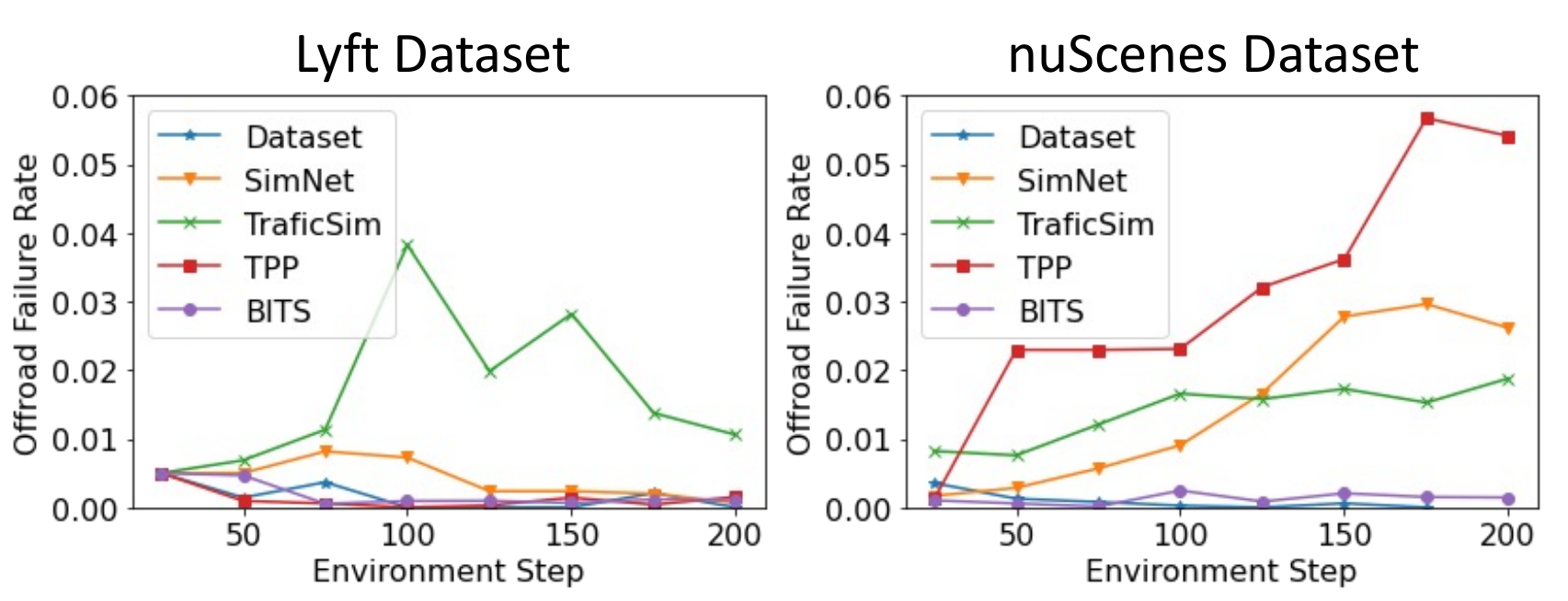}
    \hspace{0.5cm}
    \includegraphics[width=0.3\linewidth]{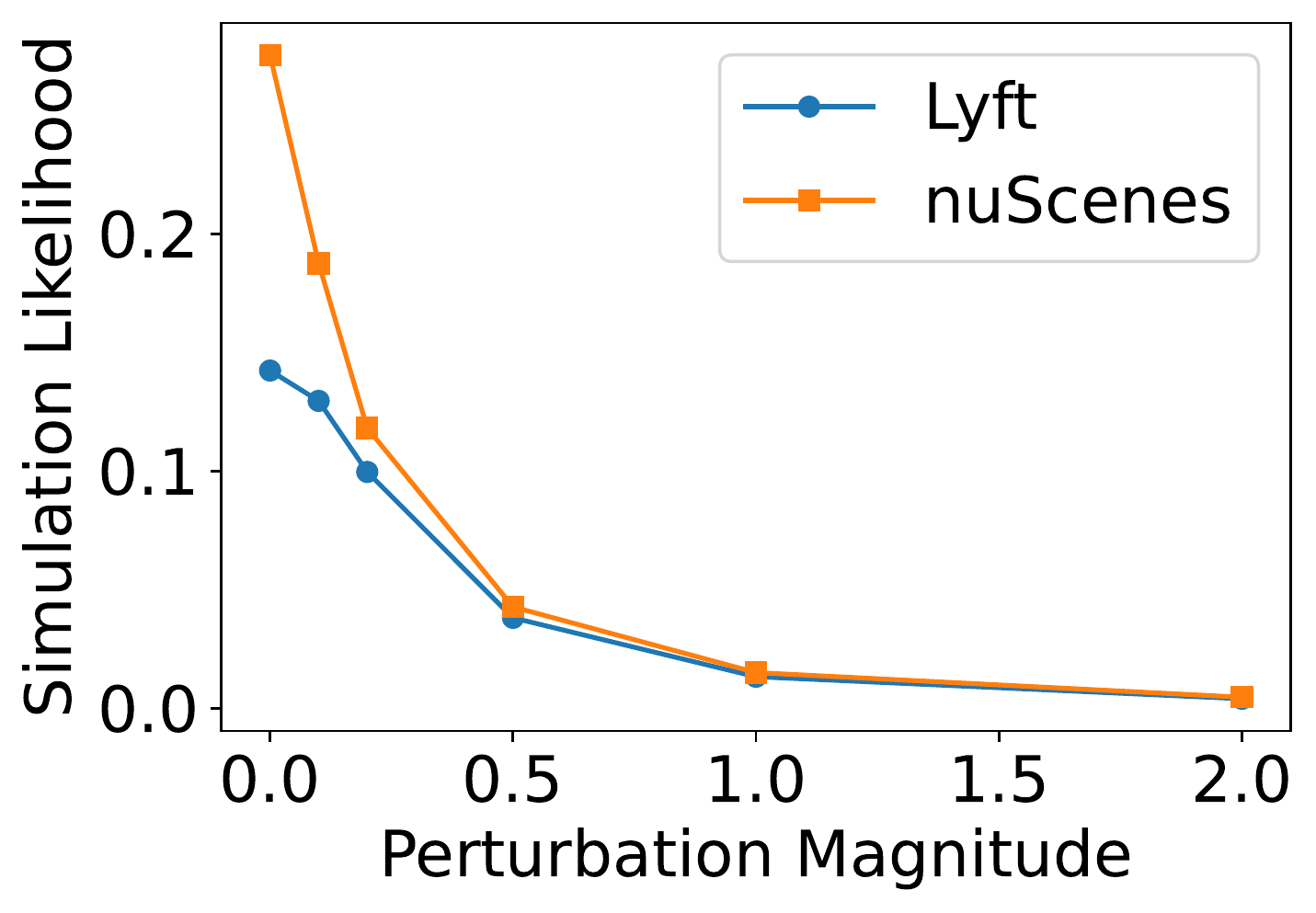}
    \vspace{-0.25cm}
    \caption{\textbf{Left:} Time-to-failure rates caused by road departure (offroad) errors. \textbf{Right:} Learned likelihood score of recorded trajectories under different levels of perturbations.}
    \label{fig:ttf}
    \label{fig:pert_GT}
    \vspace{-0.5cm}
\end{figure*}

Table~\ref{tab:lyft} and Table~\ref{tab:nusc} show quantitative results of closed-loop simulation on Lyft and nuScenes datasets, respectively. Fig.~\ref{fig:diversity} qualitatively visualizes trajectories generated by selected methods and Fig.~\ref{fig:ttf} shows a more detailed analysis on time-to-failure caused by road departure error. 
We make the following core observations from these results.

\textbf{Recorded driving data is noisy.} As stated above, both datasets contain certain levels of labeling noise indicated by non-zero failure rates for ground truth data (labeled ``Dataset"), with higher noise in Lyft than nuScenes and a majority of errors stemming from vehicle-vehicle collisions due to imprecise bounding box labels. Our method \Ours is able to achieve lower failure rates than even the recorded trajectories in both datasets.

\textbf{Sample efficiency.} nuScenes contains far fewer training samples than Lyft, necessitating high sample efficiency in order for a policy to manage compounding errors over long simulations. As shown in Table~\ref{tab:nusc}, even without the prediction-and-planning module, both variants of \Ours achieve low failure rates compared to other baselines. For a more direct analysis, we also report the mean time-to-failure in Fig.~\ref{fig:ttf}, observing that failure rates in nuScenes increase significantly over the course of simulation for the non-hierarchical policy baselines and remain low for \Ours and its ablations.

\textbf{\Ours generates diverse and stable simulations.} We observe that the baseline methods exhibit trade-offs between generating diverse rollouts and overfitting to a single mode of behaviors. For example, in Lyft, TPP suffers from mode collapse which yields low failure rates at the cost of low diversity. TrafficSim achieves relatively high diversity and coverage, but also high failure rates. This observation is corroborated by the visualizations shown in Fig.~\ref{fig:diversity}, where all simulation trials by TPP are visually identical and resembles the ground truth (on right, titled ``Dataset"), and while the trajectories generated by TrafficSim are diverse, some suffer from collisions and road departures. In contrast, \Ours simultaneously attains high diversity and coverage with a low failure rate. This contrast is more pronounced in nuScenes where the training set is small. \Ours achieves a balanced performance even without the prediction-and-planning module thanks to its high sample efficiency. 

\textbf{Prediction-and-planning is not always effective.} The prediction-and-planning module is generally effective in reducing failure rates, with two important exceptions: (1) when action samples are not diverse, and (2) when all action samples lead to failure. Case (1) is exemplified by TPP in the Lyft environment (Fig.~\ref{fig:diversity}), where the model's predictions overfit to a single behavior mode. Case (2) is exemplified by SocialGAN in both datasets. While the simulations are relatively diverse, they have high failure rates, entailing poor action sample quality. The prediction-and-planning module has negligible effects on the simulation in both cases. 



\textbf{Quantifying behavioral realism.} As discussed above, evaluating simulation realism remains a challenging open problem because there is no single correct answer for traffic simulation. Here we consider both dataset statistics and learned metrics as a proxy for quantifying behavioral realism. For Lyft, we see that all methods achieve comparable speed and jerk statistical distances relative to the recorded trajectories. As expected, SimNet has the lowest sADE due to its behavior cloning objective. In nuScenes, \Ours achieves comparable performance to SimNet in dataset metrics, showing that our method does not have to sacrifice behavioral realism for diversity and stability.

Finally, we consider the learned metric described in Sec.~\ref{ssec:metrics}. To show that this occupancy likelihood-based metric indeed captures meaningful data likelihoods, we roll out ground truth trajectories with different levels of Ornstein-Uhlenbeck noise~\cite{uhlenbeck1930theory} and measure the predicted likelihood score. As shown in Fig.~\ref{fig:pert_GT}, the likelihood score decreases smoothly as the noise intensity grows, indicating that the learned metric captures the effect of disturbances well. We report the learned likelihood scores for representative baselines in Table.~\ref{tab:likelihood}. Sensibly, we see that in both datasets the ground truth dataset trajectories have the highest likelihood scores. \Ours yields comparable or higher likelihood scores than other baselines, with scores on par with ground truth dataset trajectories on nuScenes data.

\begin{wraptable}{r}{5.5cm}

\caption{Learned likelihood scores for different traffic simulation methods.}
\vspace{1mm}
\begin{tabular}{l|cc}
\hline
                                     & Lyft  & nuScenes \\ \hline
SimNet~\cite{bergamini2021simnet}   & 0.112 & 0.167    \\
TPP~\cite{SalzmannIvanovicEtAl2020} & 0.135 & 0.155    \\
TrafficSim~\cite{suo2021trafficsim} & 0.126 & 0.176    \\
\Ours                              & 0.131 & \textbf{0.276}    \\ \hline
Dataset                              & \textbf{0.142} & \textbf{0.275}    \\ \hline
\end{tabular}
\label{tab:likelihood}
\end{wraptable}




\vspace{-0.25cm}

\section{Discussion and conclusions}

\textbf{Limitations and broader impact.} Our work has a few important limitations. First, despite our efforts, devising evaluation metrics for traffic simulation remains an open research problem. The proposed metrics can only serve as proxies for measuring behavior realism. In particular, the learned metric is likely biased by the model choice and the training data. Second, we do not consider traffic rules (e.g., driving on the correct side of the road, obeying traffic lights) in evaluation and will work on enriching our simulation software framework with additional environment constraints.
Finally, a limitation that might have broader impact is that data-driven simulation models are inherently biased by their training data, which is often curated from a small number of geographic regions. As a result, verification pipelines built on top of such models may be limited by the scenarios that they can generate. This may cause potential safety concerns for deploying tested vehicles in regions that are less represented in the training data.

\textbf{Conclusions.} In this work, we present Bi-level Imitation for Traffic Simulation (\Ours), a novel data-driven traffic simulation model. \Ours achieves high sample efficiency and behavioral diversity through a bi-level imitation learning formulation, generating stable long-horizon rollouts aided by a prediction-and-planning module. To facilitate evaluation and future studies in the field, we develop and open source a software tool that unifies data formats from different AV datasets and transforms scenes from existing datasets into interactive simulation environments. We compare \Ours against a number of competitive baselines on two large-scale real-world AV datasets and find that \Ours can generate diverse, realistic, and stable traffic simulations.

{
\small
\bibliographystyle{ieeetr}
\bibliography{main,ASL_papers,added}
}

\section{Appendix}
\begin{figure}[h!]
    \centering
    \includegraphics[width=\linewidth]{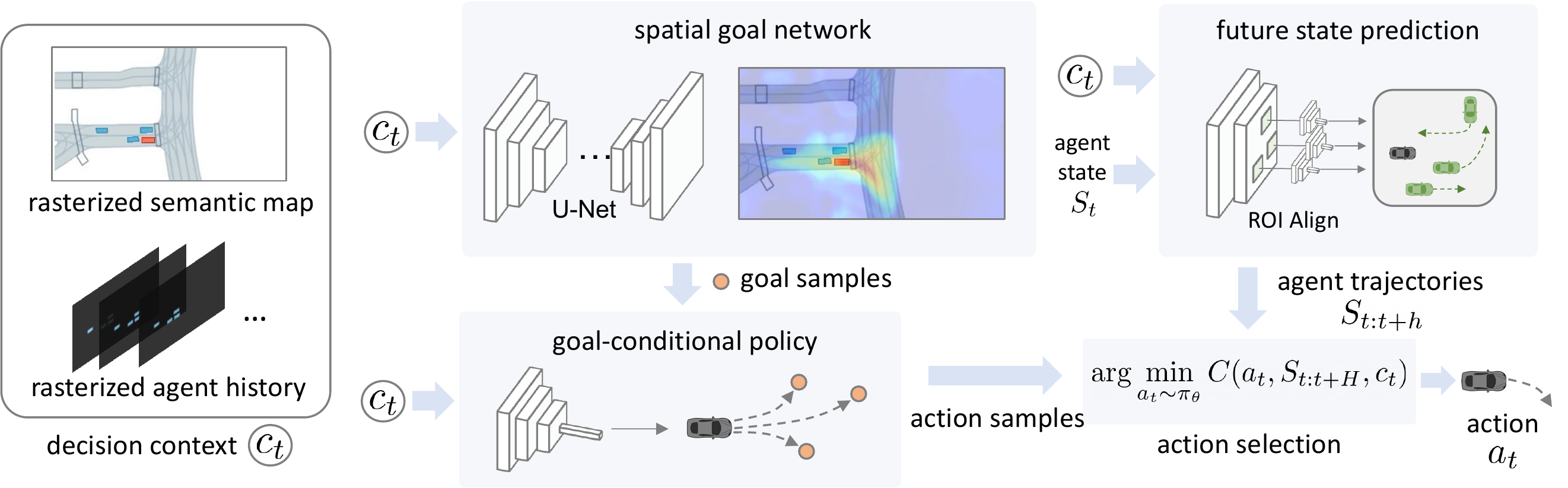}
    \caption{(Same as Fig. 1 in the main text) \textbf{\Ours framework overview}: Decision context $c_t$ is a tensor containing the semantic map and rasterized agent history concatenated channel-wise. Given $c_t$ as input, (1) the spatial goal network produces a 2D spatial distribution of short-horizon goals, (2) the goal-conditional policy generates a sequence of actions for each sampled goal, (3) a trajectory forecasting model predicts the future motion of the neighboring agents, and finally (4), based on the predicted future states, the framework selects the sequence of actions that minimizes a rule-based cost function.}
    \label{fig:model_appendix}
\end{figure}

\subsection{Implementation details}
As shown in Fig.~\ref{fig:model_appendix}, our work includes three learned component: spatial goal network, goal-conditional policy, and the future state predictor. We will describe their shared input, network backbones, and the output heads below. Table~\ref{tab:param} includes key hyperparameter selections.

\textbf{Input.} All three networks take rasterized decision context as input. A decision context $c_t$ includes (1) a semantic map including information such as lane marks, crosswalks, and lane boundaries and (2) kinematic history states of all agents (vehicles) in the view. For the semantic map, we follow the original rasterization scheme included in Lyft~\cite{HoustonZuidhofEtAl2020} and nuScenes~\cite{CaesarBankitiEtAl2019} development kit, respectively. Specifically, Lyft semantic map is a 3-channel tensor that includes lane boundary, lane areas, traffic light faces, and crosswalk information. nuScenes's semantic maps are 7-channel tensors including lane, road segment, drivable area, road divider, lane divider, pedestrian crossing, and walkaway. Following prior works~\cite{bergamini2021simnet,konev2021motioncnn}, we use a binary occupancy map same size as the semantic map to represent the kinematic states (i.e., position, heading, and extent) of agents at a given timestep. Each agent in the view is rasterized as filled bounding boxes on the occupancy map. State histories are represented as multi-channel occupancy maps. We concatenate the semantic map and the rasterized agent history map channel-wise and use the tensor as input for all models.

\textbf{Architecture details.} All three networks use an identical ResNet-18 ConvNet backbone to encode an input decision context tensor into a compact feature vector. Here we describe the output heads of each network.
\begin{table}[]
\centering
\caption{Key parameters for \Ours}
\begin{tabular}{l|c}
\hline
Key params                   & Values         \\ \hline
Step time (s)                & 0.1           \\
History length (step)        & 10            \\
Prediction length (step)     & 20            \\ \hline
Input                        & Values         \\ \hline
Semantic map size - Lyft     & (3, 224, 224) \\
Semantic map size - nuScenes & (7, 224, 224) \\
Pixel size (m/pixel)         & 0.5           \\ \hline
Training params              & Values         \\ \hline
Learning rate                & 0.0001        \\
Batch size                   & 100           \\
Optimizer                    & Adam         \\
Loss - trajectory prediction & L2 regression \\
Loss - spatial occupancy     & CrossEntropy  \\ \hline
Simulation params            & Values         \\ \hline
Num simulation steps         & 200           \\
Num action samples           & 50            \\
$n$-step action              & 5             \\
Planning cost weight - collision             & 10.0          \\
Planning cost weight - offroad               & 1.0           \\ \hline
\end{tabular}
\label{tab:param}
\end{table}

\emph{Goal-conditional policy.} Given the feature and a goal pose, i.e., target position and heading in the agent coordinate frame, the goal-conditional policy uses a MLP-based trajetory decoder to generate a length-$h$ trajectory as actions. As mentioned in the main text, inspired by prior works~\cite{SalzmannIvanovicEtAl2020,suo2021trafficsim}, instead of directly regressing each state in an agent's trajectory, the decoder predicts control inputs (velocity, change of heading) at each future time step and forward integrates them through an agent's dynamics model (e.g., extended unicycle dynamics~\cite{LaValle2006BetterUnicycle} for vehicles) to obtain the trajectory predictions. The model is trained end-to-end with an L2 regression loss against the recorded trajectories in the dataset, similar to a regular behavior cloning objective. 

\emph{Future state predictor.} The network is part of our prediction-and-planning module that selects action based on the predicted future states of the environment. Since we assume an analytical vehicle dynamics model and known static map, the main task of the network is to predict the future motion trajectories of nearby agents. As described in the main text, we follow a typical trajectory prediction pipeline and featurize each agent by its local and global scene context. Specifically, we use RoIAlign~\cite{he2017mask} to crop the features extracted at an intermediate layer of the ResNet-18 encoder. Specifically, we use a cropping window of $7 \times 7$ at the end of the second block of the network. The per-agent features are then concatenated with a global scene context feature (extracted by the final fully-connected layer of the ResNet-18) to make the final trajectory prediction $s^{(i)}_{t:t+H}$ for each neighboring agent $i$. We only make predictions for agents that are visible in the rasterized map. The model is trained with an L2 regression loss against the recorded neighboring agent trajectories in the dataset. The goal-conditional policy and the future state prediction network share encoder network weights and are trained jointly.

\emph{Spatial goal network.}  The spatial goal predictor extends the ConvNet encoder backbone with a U-Net-style decoder network (mirroring the encoder size) to generate the spatial goal prediction. Specifically, the model generates a 2D grid of goal likelihood as well as residual components to refine the predicted goal location. The output takes the shape of a 4-channel tensor with the same spatial size as the input rasterized map. Channel $0$ is the likelihood of the coarse goal location 2D probability map. We use a grid size of 0.5 meter per pixel. Each pixel in channel $1$ and $2$ is the $(x, y)$ scalar residual (in meters) relative to the grid location. Channel $3$ is the heading prediction at each grid location in radian. Once a location is selected based on the probability map in channel $0$, the location is corrected by the residual and transformed into a goal pose $\tilde{s}_{t+H}$ in the agent local coordinate frame. Note that it is possible to further discretize the heading prediction into discrete bins~\cite{zeng2018learning}, but we empirically found that the heading prediction has negligible impact on the goal-conditional policy. We treat the 2D location map as a joint distribution and train via cross-entropy loss across locations. The other channels are trained with regression losses (e.g., squared error). 

\textbf{Training setup.} We train all models on shared cloud computing nodes equipped with NVIDIA Tesla V100 GPUs (32gb GPU memory). We train all models for 100k iterations (gradient steps) and choose the checkpoint based on their respective offline validation metrics (e.g., prediction error). On average, training takes between 20-30 hours to complete with single-GPU jobs. 

\begin{figure}
    \centering
    \includegraphics[width=0.4\columnwidth]{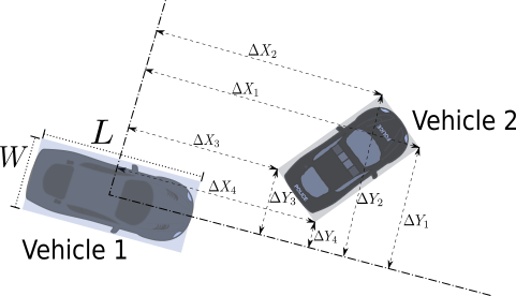}
    \caption{Vehicle-vehicle collision check illustration.}
    \label{fig:vv_col}
\end{figure}

\textbf{Cost functions for prediction and planning.} As mentioned in the main text, we consider two rule-based costs: collision and road departure. We follow prior work~\cite{SalzmannIvanovicEtAl2020} and compute the collision cost based on distances from the corners of two bounding box rectangles using distances between the four corners. The minimum distance $d_{\min}(\cdot)$ is approximated as:
\begin{equation*}
    d_{\min}(\Delta X_{1:4},\Delta Y_{1:4},L,W) =\max\left\{\begin{aligned}|&\Delta X_1|-\frac{L}{2},...,|\Delta X_4|-\frac{L}{2},\\&|\Delta Y_1|-\frac{W}{2},...,|\Delta Y_4|-\frac{L}{2}\end{aligned}\right\},
\end{equation*}
which is further illustrated in Fig.~\ref{fig:vv_col}. We then add the following collision loss
\begin{equation*}
    \text{Col}=\text{Sigmoid}(-\alpha d_{\min}-\beta),
\end{equation*}
where $\alpha$ and $\beta$ are parameters used to shape the sigmoid loss. In our experiments, $\alpha=1.0,~\beta=4.0$.

To calculate the road departure cost, we first generate a distance map that records the Manhattan distance to the drivable area in pixel space. To be specific, we first pick a maximum distance constant $D$ and set all pixels inside the drivable area to 0, and all pixels outside the drivable area to $D$. Then we perform the following convolution step $D$ times:
\begin{equation*}
    x_{i,j}=\min\{x_{i,j},x_{i-1,j}+1,x_{i+1,j}+1,x_{i,j-1}+1,x_{i,j+1}+1\},
\end{equation*}
where $x_{ij}$ is the value at the $i,j$ coordinate. The resulting distance map assigns zero to points within the drivable area, with values increasing outside the drivable area until saturating at $D$. We directly use this Manhattan distance value as a penalty. To account for the size of the vehicle, we perform RoIAlign on the distance map with vehicle patches that take the vehicles' size and orientation into account. Note that both cost terms are close to zero (the collision loss may be nonzero due to the sigmoid function) for nominal trajectories, i.e., trajectories that do not result in collision and road departure, minimizing their effect on selecting among rule-following action samples. 

\textbf{Learned metrics.} To compute likelihood of a simulated trajectory, we use an occupancy-based prediction model that predicts where an agent will be in future time steps. The model uses a similar structure to our spatial goal network and discretizes the position space into bins. The model is trained to minimize a cross-entropy loss function and its predictions are then used to compute the trajectory likelihoods of simulated trajectories.  During evaluation, the likelihood of an agent's trajectory is calculated in a receding horizon manner. For each time step, we compare its rollout trajectory for the next $T=20$ steps to the predicted occupancy and calculate the likelihood and aggregate the whole rollout duration by taking the mean. The likelihood values of multiple agents in the scene are also aggregated by taking the mean.



\subsection{Baselines}
\textbf{SimNet.} SimNet~\cite{bergamini2021simnet} is a deterministic behavior-cloning model for traffic simulation. Prior work also show that its rasterized encoding scheme yields strong performance in trajectory prediction tasks~\cite{konev2021motioncnn}. We use the same rasterization input and CNN-based encoding backbone for all baselines and our \Ours model.

\textbf{SocialGAN.} SocialGAN is adapted from a trajectory prediction model~\cite{GuptaJohnsonEtAl2018}. The model is trained using conditional generative-adversarial learning, where the condition is the decision context and the generation target comes from the trajectory dataset. To facilitate a fair comparison, the model uses the same encoding backbone and MLP-based trajectory decoder as all other methods. 

\textbf{TPP (Trajectron++).} TPP is adapted from the Trajectron++ model~\cite{SalzmannIvanovicEtAl2020}, which is a CVAE model with a discrete latent space and generates a Gaussian Mixure Model (GMM) as trajectory prediction. We modified the encoder to take rasterized features as input, which shares the same map encoder as the proposed \Ours model. The TPP model generates predictions of the control input, which are propagated through a unicycle model to obtain the trajectory prediction. For simplicity, we directly let the decoder predict the variance of predicted trajectory instead of propagating the variance through the dynamics as in \cite{SalzmannIvanovicEtAl2020}. The latent dimension is chosen to be 10 in our experiments.

\textbf{TrafficSim.} TrafficSim is an agent-centric adaptation of the original TrafficSim work~\cite{suo2021trafficsim}, meaning that agents in a scene are controlled by independent replica of the same model instead of by a single model that explicitly coordinates their actions. The TrafficSim baseline uses an isotropic Gaussian CVAE to model action distributions and facilitate diverse simulation. We adopt a similar differentiable dynamics trajectory decoder from the original paper but remove the scene consistency loss in training since we do not assume control over all agents.

\section{Additional results}
\subsection{Complete evaluation metrics}
Here we provide the exhaustive list of (non-learned) evaluation metrics we considered in this work and report performances in Table~\ref{tab:lyft_complete} for Lyft and Table~\ref{tab:nusc_complete} for nuScenes.
\begin{itemize}
    \item \textbf{FR}: The average fraction of failed agent in each scene. The failure can be caused by either collision failure or offroad failure (defined below).
    \item \textbf{collFR}: The average fraction of failed agent in each scene, where the failure is caused by agent colliding with any other agent in the scene at any timestep throughout an episode. 
    \item \textbf{offroadFR}: The average fraction of failed agent in each scene, where the failure is caused by agent departing drivable region for more than 1 second (10 timesteps for a simulation frequency of 0.1 second). 
    \item \textbf{coll}: The average fraction of agent colliding with another agent. We in addition list the type of collision occurred (rea, front, side). Note that each collision type is independently calculated, i.e., each agent can have multiple type of collision per episode.
    \item \textbf{offroad}: The fraction of timesteps that an agent spend outside of drivable area, averaged across agents per scene.
    \item \textbf{coverage} and \textbf{diversity}: As described in detail in the main text, both metrics measure the diversity of rollouts across multiple simulation trials with the same initial condition. The coverage metric calculates the (discretized) map area covered by multiple simulation trials with the same initial condition. The diversity metric in addition computes the Wasserstein distances among the density profiles of the map coverage attained by different simulation trials. In practice, we accumulate the metrics across 5 trials and compute the coverage density by discretizing the map at a 2m by 2m resolution.
    \item \textbf{Dataset metrics}. Dataset metrics compare the driving profile of simulation trajectories with trajectories recorded in the dataset. We consider \textbf{speed}, \textbf{lon acc} (longitudinal acceleration magnitude), \textbf{lat acc} (latitudinal acceleration magnitude), and \textbf{jerk}. To calculate the difference between the driving profile of the simulation and the dataset, we first collect histograms of these driving profile measurements for both the simulated and the dataset trajectories across all evaluation scenes. We then calculate the distances between the histograms of the simulation and the dataset using Wasserstein distance normalized by a constant. We use 20 bins for all histograms, with a speed range of $[0, 30](m/s)$, an acceleration magnitude range of $[0, 10] (m/s^2)$, and a jerk magnitude range of $[0, 10](m/s^3)$.
    
    We also report scene Average Distance Error (\textbf{sADE}) and scene Final Distance Error (\textbf{sFDE}), which measures the average and the final position differences between simulated and recorded trajectories, respectively. Note that sADE and sFDE are not suitable for measuring simulation realism and is included only as a reference since it heavily penalizes alternative simulations (e.g., turning left vs.~going straight). 
\end{itemize}

\begin{table}[]
\small
\caption{Complete results on the Lyft dataset.}
\begin{tabular}{l|cccccccc}
\hline
                                       & FR       & collFR   & offroadFR & coll (any) & coll (rear) & coll (front) & coll (side) & offroad \\ \hline
SimNet\cite{bergamini2021simnet}     & 38.35    & 35.57     & 3.61       & 35.57      & 20.58       & 20.46        & 30.59       & 1.38    \\
SocialGAN\cite{GuptaJohnsonEtAl2018} & 64.96    & 42.86     & 41.00      & 42.86      & 16.26       & 15.78        & 39.60       & 19.45   \\
SocialGAN+p                            & 69.41    & 42.96     & 48.02      & 42.96      & 15.56       & 15.26        & 40.39       & 25.12   \\
TPP\cite{SalzmannIvanovicEtAl2020}   & 15.62    & 14.65     & 0.98       & 14.65      & 7.23        & 6.94         & 8.00        & 0.59    \\
TPP+p                                  & 16.03    & 15.12     & 1.02       & 15.12      & 7.41        & 7.14         & 10.85       & 0.65    \\
TrafficSim\cite{suo2021trafficsim}   & 26.98    & 15.98     & 13.53      & 15.98      & 6.43        & 6.58         & 8.59        & 5.76    \\
TrafficSim+p                           & 22.97    & 13.58     & 11.39      & 13.58      & 5.90        & 5.82         & 7.23        & 4.70    \\
\Ours (max)                          & 20.71    & 18.75     & 2.36       & 18.75      & 7.44        & 7.08         & 12.72       & 1.18    \\
\Ours (sample)                       & 25.37    & 22.37     & 3.94       & 22.37      & 9.62        & 9.93         & 15.60       & 1.27    \\
\Ours                                & 9.97     & 8.66      & 1.48       & 8.66       & 3.64        & 3.61         & 6.79        & 0.46    \\ \hline
Dataset                                & 18.36    & 17.49     & 1.22       & 17.49      & 6.14        & 7.31         & 14.59       & 0.84    \\ \hline
                                       & coverage & diversity & speed      & lon acc    & lat acc     & jerk         & sADE        & sFDE    \\ \hline
SimNet\cite{bergamini2021simnet}     & 460.44   & 0.00      & 1.65       & 19.16      & 21.26       & 3.85         & 3.19        & 8.43    \\
SocialGAN\cite{GuptaJohnsonEtAl2018} & 189.98   & 9.02      & 1.48       & 10.10      & 16.21       & 5.17         & 13.22       & 32.74   \\
SocialGAN+p                            & 131.47   & 7.64      & 1.51       & 9.66       & 15.53       & 4.78         & 13.90       & 34.09   \\
TPP\cite{SalzmannIvanovicEtAl2020}   & 495.69   & 3.23      & 1.14       & 17.42      & 19.09       & 2.60         & 5.50        & 11.85   \\
TPP+p                                  & 508.16   & 2.75      & 1.23       & 17.93      & 19.72       & 2.32         & 5.44        & 11.75   \\
TrafficSim\cite{suo2021trafficsim}   & 566.35   & 7.68      & 1.50       & 17.20      & 21.42       & 3.82         & 5.89        & 14.02   \\
TrafficSim+p                           & 617.50   & 7.96      & 1.73       & 17.84      & 21.74       & 3.09         & 6.04        & 13.94   \\
\Ours (max)                          & 443.93   & 0.00      & 0.76       & 15.10      & 19.18       & 4.44         & 7.77        & 16.74   \\
\Ours (sample)                       & 780.52   & 16.84     & 1.86       & 18.20      & 20.38       & 4.29         & 7.78        & 17.25   \\
\Ours                               & 1014.43  & 22.94     & 1.96       & 17.05      & 20.76       & 3.75         & 11.21       & 23.05   \\ \hline
Dataset                                & 327.25   & 0.00      & 0.00       & 0.00       & 0.00        & 0.00         & 0.00        & 0.00    \\ \hline
\end{tabular}
\label{tab:lyft_complete}
\end{table}

\begin{table}[]
\small
\caption{Complete results on the nuScenes dataset.}
\begin{tabular}{l|cccccccc}
\hline
                                       & FR       & coll FR   & offroad FR & coll (any) & coll (rear) & coll (front) & coll (side) & offroad \\ \hline
SimNet\cite{bergamini2021simnet}     & 24.58    & 15.80     & 11.95      & 15.80      & 5.91        & 7.02         & 13.17       & 3.05    \\
SocialGAN\cite{GuptaJohnsonEtAl2018} & 71.33    & 33.81     & 51.24      & 33.81      & 22.27       & 23.86        & 30.41       & 19.71   \\
SocialGAN+p                            & 71.86    & 35.84     & 51.36      & 35.84      & 23.54       & 25.26        & 30.51       & 18.26   \\
TPP\cite{SalzmannIvanovicEtAl2020}   & 49.76    & 33.23     & 24.93      & 33.23      & 13.48       & 13.85        & 28.88       & 8.27    \\
TPP+p                                  & 9.84     & 9.52      & 0.32       & 9.52       & 2.38        & 2.38         & 7.97        & 0.04    \\
TrafficSim\cite{suo2021trafficsim}   & 27.08    & 20.39     & 11.16      & 20.39      & 7.60        & 8.29         & 17.48       & 3.42    \\
TrafficSim+p                           & 25.97    & 18.45     & 12.01      & 18.45      & 6.68        & 6.91         & 16.60       & 3.61    \\
\Ours (max)                          & 13.24    & 11.51     & 1.83       & 11.51      & 4.11        & 4.24         & 8.31        & 0.30    \\
\Ours (sample)                       & 14.72    & 13.79     & 1.20       & 13.79      & 4.98        & 5.28         & 10.47       & 0.17    \\
\Ours                                & 6.63     & 5.67      & 1.01       & 5.67       & 1.85        & 1.91         & 3.87        & 0.14    \\ \hline
Dataset                                & 11.99    & 11.39     & 0.64       & 11.39      & 4.72        & 6.44         & 9.26        & 0.42    \\ \hline
                                       & coverage & diversity & speed      & lon acc    & lat acc     & jerk         & sADE        & sFDE    \\ \hline
SimNet\cite{bergamini2021simnet}     & 395.21   & 0.00      & 6.81       & 144.06     & 125.14      & 18.61        & 7.01        & 17.87   \\
SocialGAN\cite{GuptaJohnsonEtAl2018} & 154.21   & 2.29      & 38.44      & 319.37     & 342.78      & 42.51        & 19.89       & 36.79   \\
SocialGAN+p                            & 151.61   & 2.32      & 37.07      & 316.63     & 339.96      & 42.48        & 19.76       & 36.68   \\
TPP\cite{SalzmannIvanovicEtAl2020}   & 489.36   & 7.06      & 8.37       & 137.54     & 117.77      & 4.61         & 12.13       & 27.26   \\
TPP+p                                  & 661.07   & 7.73      & 8.17       & 138.76     & 114.82      & 5.96         & 13.09       & 28.65   \\
TrafficSim\cite{suo2021trafficsim}   & 861.00   & 4.28      & 10.03      & 176.83     & 156.61      & 10.01        & 10.51       & 25.77   \\
TrafficSim+p                           & 933.53   & 4.43      & 10.44      & 179.63     & 159.45      & 11.93        & 11.06       & 27.39   \\
\Ours (max)                          & 559.58   & 0.00      & 6.35       & 150.67     & 124.78      & 13.87        & 6.41        & 15.71   \\
\Ours (sample)                       & 888.12   & 6.36      & 6.54       & 152.42     & 125.97      & 13.44        & 6.70        & 16.22   \\
\Ours                                & 1122.94  & 8.40      & 6.91       & 158.61     & 133.00      & 13.20        & 8.39        & 20.19   \\ \hline
Dataset                                & 397.92   & 0.00      & 0.00       & 0.00       & 0.00        & 0.00         & 0.00        & 0.00    \\ \hline
\end{tabular}
\label{tab:nusc_complete}
\end{table}
\subsection{Planning cost weight ablation} 
\begin{figure}[h]
    \centering
    \includegraphics[width=0.75\linewidth]{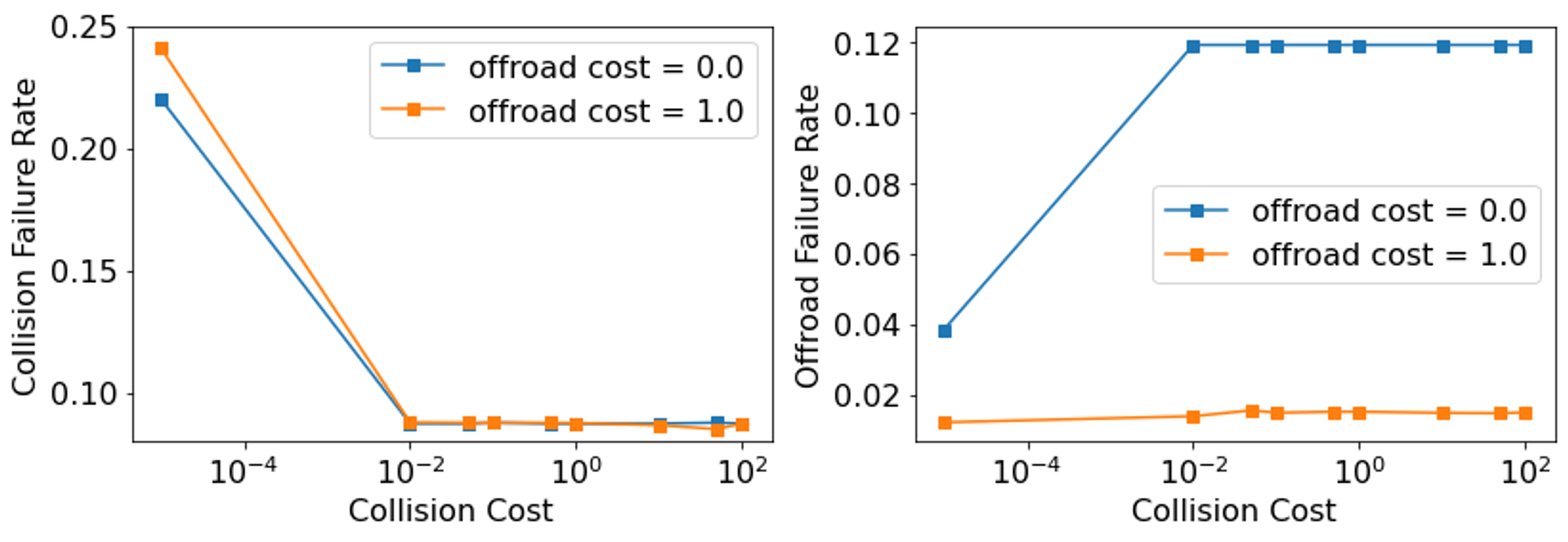}
    \caption{Failure rates of \Ours with different planning cost weights.}
    \label{fig:cost}
\end{figure}

As described above and in the main text, we use two cost terms: collision and road departure in our prediction-and-planning module. We conduct an ablation study on different planning cost weight setting using the Lyft dataset and report the result in Fig.~\ref{fig:cost}. Because there are only two cost terms, we mainly consider the ratio of the cost weights. We set offroad cost weight to either $0.0$ and $1.0$ and continuously vary the collision cost weight at a log scale. We observe that both terms are effective at reducing rates of their corresponding cause of failure, the reduction saturates when the weight is above a certain value, and the overall failure reduction performance is not sensitive to the cost weight ratio.  

\subsection{Prediction and planning horizon ablation} Here we investigate the impact of the planning horizon $H$ in \Ours's planning and control module. We train all three components (i.e., goal-conditional policy, future state predictor, spatial goal network) with the same set of prediction horizons: $\{10, 20, 50, 80\}$ on the Lyft dataset. We report performances using essential metrics in Table~\ref{tab:horizon}. We observe that $H=\{10, 20\}$ yield similar performances, but the failure rates increase significantly as the model expands its prediction horizon to $H=\{50, 80\}$. This is because predicting long-term future is extremely challenging, especially when all agents are controlled by stochastic policies. The model's decisions can be easily misled by wrong future predictions, leading to critical mistakes. At the same time, we also observe that while short-term predictions result in more successful simulation, they lead to less smooth driving behavior caused by more frequent braking and acceleration, as indicated by the dataset metrics (e.g., higher acc and jerk). Hence an important future direction is to dynamically adjust the prediction and planning horizon by taking into account prediction uncertainties. 

\begin{table}[h]
\caption{Performance of \Ours using different prediction and planning horizon $H$ on the Lyft dataset.}
\begin{tabular}{c|ccccc|cccc}
\hline
horizon $H$ & FR & coll & offroad & coverage & diversity & speed & lon acc & lat acc & jerk \\ \hline
10        & 8.89           & 6.94             & 0.75                & 948.41             & 23.31               & 1.58  & 17.50   & 20.93   & 4.03 \\
20        & 9.97           & 8.66             & 0.46                & 1014.43            & 22.94               & 1.96  & 17.05   & 20.76   & 3.75 \\
50        & 23.86          & 21.30            & 0.75                & 654.25             & 15.14               & 0.69  & 12.67   & 17.91   & 3.50 \\
80        & 32.39          & 28.62            & 1.48                & 355.69             & 11.43               & 1.83  & 9.31    & 15.46   & 3.50 \\ \hline
\end{tabular}
\label{tab:horizon}
\end{table}


\subsection{Additional qualitative results} 
\begin{figure}
    \centering
    \includegraphics[width=\linewidth]{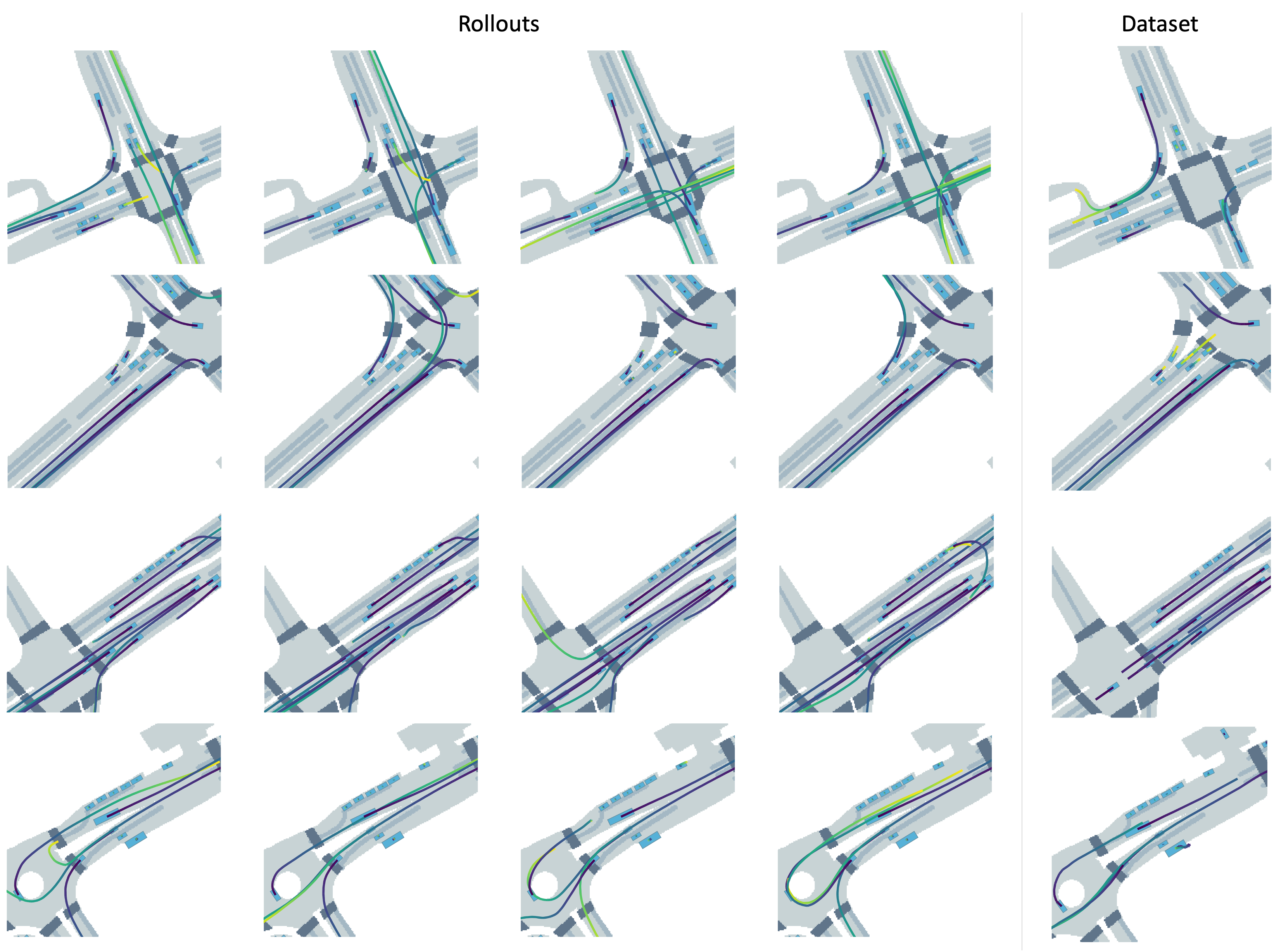}
    \caption{Qualitative results on the nuScenes dataset.}
    \label{fig:qual_nusc}
\end{figure}

\begin{figure}
    \centering
    \includegraphics[width=\linewidth]{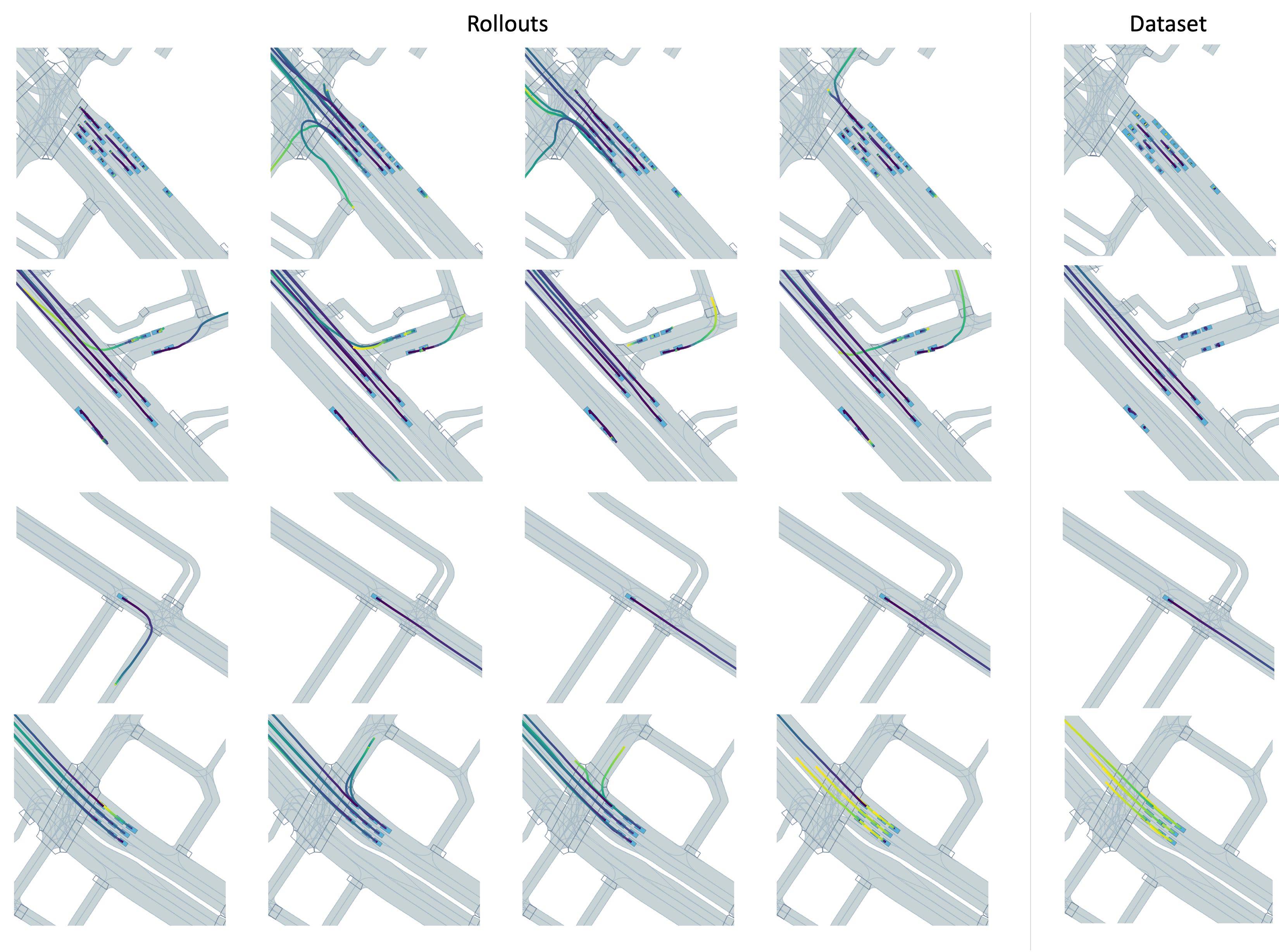}
    \caption{Qualitative results on the Lyft dataset.}
    \label{fig:qual_lyft}
\end{figure}

We include additional qualitative results for \Ours. Rollouts using the nuScenes dataset is shown in Fig.~\ref{fig:qual_nusc} and that of the Lyft dataset is shown in Fig.~\ref{fig:qual_lyft}. Each row shows different rollouts from the same initial scene configuration. The recorded trajectories are shown in the rightmost column for reference. 

\end{document}